\documentclass[final]{cvpr}

\usepackage[linesnumbered,boxed,ruled,commentsnumbered]{algorithm2e}
\usepackage{times}
\usepackage{epsfig}
\usepackage{graphicx}
\usepackage{amsmath}
\usepackage{amssymb}
\usepackage[utf8]{inputenc} 
\usepackage[T1]{fontenc}    
\usepackage{url}            
\usepackage{booktabs}       
\usepackage{amsfonts}       
\usepackage{nicefrac}       
\usepackage{microtype}      
\usepackage{graphicx}
\usepackage{multirow}
\usepackage{wrapfig}
\usepackage{xcolor}
\usepackage{bbm}
\usepackage{amsmath}
\usepackage{comment}
\usepackage{dsfont}
\usepackage{bbm}
\DeclareUnicodeCharacter{2212}{-}
\usepackage[pagebackref=true,breaklinks=true,colorlinks,bookmarks=false]{hyperref}

\def\cvprPaperID{6084} 
\def\confYear{CVPR 2021}

\begin{document}

\title{You Only Need One Thing One Click: \\Self-Training for
Weakly Supervised 3D Scene Understanding}

\author{Zhengzhe Liu$^{1}$ \quad   Xiaojuan Qi$^{2}$ \quad     Chi-Wing Fu$^{1}$ \\
$^1$The Chinese University of Hong Kong \quad $^2$The University of Hong Kong\\
{\tt\small \{zzliu,cwfu\}@cse.cuhk.edu.hk \quad  xjqi@eee.hku.edu.hk}
}

\maketitle

\begin{abstract}

3D scene understanding, e.g., point cloud semantic and instance segmentation, often requires large-scale annotated training data, but clearly, point-wise labels are too tedious to prepare.
While some recent methods propose to train a 3D network with small percentages of point labels, we take the approach to an extreme and propose ``One Thing One Click,'' meaning that the annotator only needs to label one point per object.
To leverage these extremely sparse labels in network training, we design a novel self-training approach, in which we iteratively conduct the training and label propagation, facilitated by a graph propagation module.
Also, we adopt a relation network to generate the per-category prototype to enhance the pseudo label quality and guide the iterative training. Besides, our model can be compatible to 3D instance segmentation equipped with a point-clustering strategy. 

Experimental results on both ScanNet-v2 and S3DIS show that our self-training approach, with extremely-sparse annotations, outperforms all existing weakly supervised methods for 3D semantic and instance segmentation by a large margin, and our results are also comparable to those of the fully supervised counterparts. Codes and models are available at {\scriptsize \url{https://github.com/liuzhengzhe/One-Thing-One-Click}}.  
\end{abstract}

\section{Introduction}

\begin{figure}
\centering
\includegraphics[width=0.99\columnwidth]{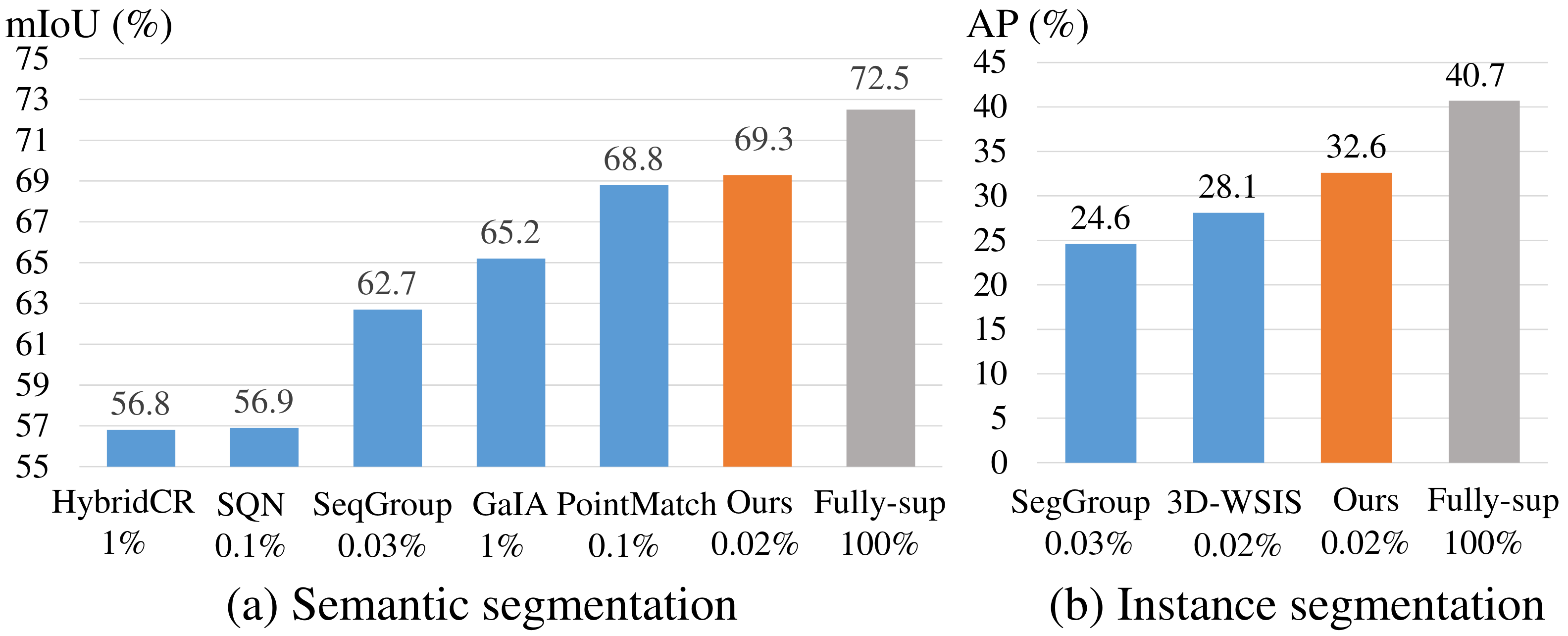}
\caption{Comparing our approach of ``One Thing One Click++'' and the fully supervised version of our method ``Fully-sup'' on 3D semantic and instance segmentation of ScanNet-v2. It is worth noting that all the works included in this comparison in the blue color were proposed recently in either 2022 or 2023. 
Our approach outperformed these very recent works by {\em training on data with only one label per object\/}.
Note the annotation percentages under each method in the charts.
}\label{fig:illustration}
\end{figure}

\begin{figure}
\centering
\includegraphics[width=0.99\columnwidth]{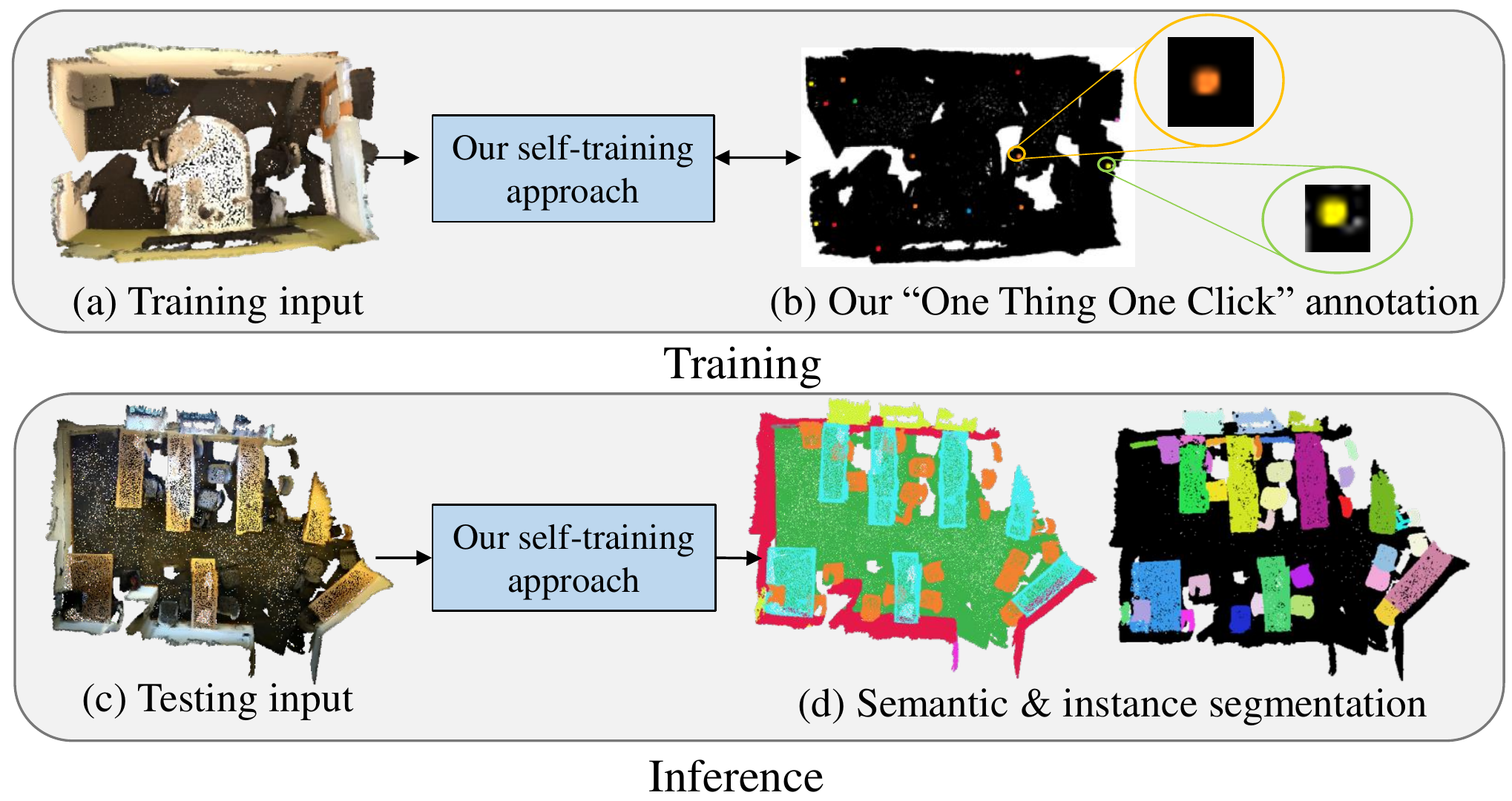}
\caption{We train our self-training approach using only our ``One Thing One Click'' annotations (top).
Yet, it can produce plausible semantic and instance segmentation results. } 
 \label{fig:fig2}

\end{figure}

The success of 3D semantic segmentation benefits a lot from the large annotated training data. However, annotating a large amount of point cloud data is exhausting and costly. Taking ScanNet-v2\cite{dai2017scannet} as an example, it takes 22.3 minutes to annotate one scene on average. It is a great burden to annotate the whole data set, which includes 1,513 scenes, thus potentially restricting further applications that require larger scale data.
Thus, efficient approaches to facilitate 3D data annotation are highly desirable.

Recently, some methods~\cite{wei2020multi,wang2020weakly,xu2020weakly} were proposed to reduce efforts to annotating 3D point clouds. Though they improve annotation efficiency, various issues remain.
Scene-level annotation in~\cite{wei2020multi} could impose negative effects on the model in the absence of localization information, whereas sub-cloud annotation in~\cite{wei2020multi} requires an extra burden to first divide the input into subclouds and then repeatedly annotate semantic categories in individual subclouds.
The 2D image annotation approach~\cite{wang2020weakly} requires extra labor to prepare a 2D image annotation, which is also a tedious task on its own.
Xu~\etal~\cite{xu2020weakly} presume that the labeled points follow a uniform distribution.
Such a requirement can be achieved by subsampling from a fully-annotated dataset, but is hard for the annotators to follow in practice. Very recent works~\cite{li2022hybridcr,hu2022sqn,lee2023gaia,wu2022pointmatch} further improve the performance with fewer annotations; yet, there is still a certain performance gap compared with the fully-supervised approaches; see Figure~\ref{fig:illustration}. 

In this work, we also aim to reduce the amount of necessary annotations on point clouds, but we take the approach to an extreme by proposing ``One Thing One Click,'' so the annotator only needs to label one single point per object.
To further relieve the annotation burden, such a point can be randomly chosen, not necessarily at the center of the object.
On average, it takes less than 2 minutes to annotate a ScanNet-v2 scene with our ``One Thing One Click'' scheme (see an example annotation in Figure~\ref{fig:fig2} (b), which contains only 13 clicks), which is more than 10x faster compared with the original ScanNet-v2 annotation scheme.

However, directly training a network on extremely-sparse labels from our annotation scheme (less than 0.02\% in ScanNet-v2 and S3DIS) will easily make the network overfit the limited data and restrict its generalization ability.
Hence, it raises the question ``can we achieve a performance comparable to a fully supervised baseline given the extremely-sparse annotations?''

To meet such a challenge, we propose to design a self-training approach with a label-propagation mechanism for weakly supervised semantic segmentation.
On the one hand, with the prediction result of the model, the pseudo labels can be expanded to unknown regions through our graph propagation module.
On the other hand, with richer and higher quality labels being generated, the model performance can be further improved.
Thus, we conduct the label propagation and network training iteratively, forming a closed loop to boost the performance of each other.

A core problem of label propagation is how to measure the similarity among nodes, especially on irregular 3D point clouds. Previous works on 2D image recognition ~\cite{zheng2015conditional,chen2017deeplab,yuan2019structpool} build a graph model upon 2D pixels and measure the similarity with low-level image features, e.g., coordinates and colors.
On the contrary, we propose a graph model building upon the 3D geometric coherent super-voxels, which have more complex geometric structures and a variable number of points in each group. Hence, existing hand-craft features cannot fully reveal the similarity among nodes in our case.
To resolve this problem, we further propose a relation network to leverage 3D geometrical information for similarity learning among the graph nodes in 3D. The geometrical similarity and learned similarity are integrated together to facilitate label propagation. To effectively train the relation network with the extremely-sparse and category-unbalanced data, we further propose to generate a category-wise prototype with a memory bank for better similarity measurement.


Further, our approach is ready for 3D instance segmentation with a point-clustering strategy.  
Leveraging the knowledge of the number and location of each instance provided by ``One Thing One Click'' annotation, point clustering aims to group the points of the same instance to generate instance-level pseudo label and enable the instance-level understanding. 

Experiments conducted on two public data sets ScanNet-v2 and S3DIS manifest the effectiveness of the proposed method.
With just around 0.02\% point annotations, our approach 
achieves results that are comparable with a fully supervised counterpart; see Figure~\ref{fig:illustration}. These results manifest the high efficiency of our ``One Thing One Click'' scheme for 3D point cloud annotation and the effectiveness of our self-training approach for weakly supervised 3D scene understanding.

This work extends our research work presented in ``One Thing One Click''~\cite{liu2021one} , which was presented at the 2021 IEEE Conference on Computer Vision and Pattern Recognition (CVPR). First, we expand the self-training framework for weakly-supervised 3D instance segmentation. Besides, we provide an alternative design for the Relation Network, which achieves comparable performance to the original design in ``One Thing One Click'' but is more efficient. At last, we compare our approach with the latest methods on 3D semantic and instance segmentation, and the experimental results demonstrate the superiority of our new approach.

\section{Related Work}

\paragraph{Semantic Segmentation for Point Cloud}
Approaches for 3D semantic segmentation can be roughly divided into point-based methods and voxel-based methods.

\textit{Point-based networks} take raw point clouds as input.  Along this line of works, PointNet~\cite{qi2017pointnet} and PointNet++~\cite{qi2017pointnet++} are the pioneering ones. 
Afterward, convolution-based methods~\cite{li2018pointcnn,thomas2019kpconv,wu2019pointconv,boulch2020convpoint} were also proposed for 3D semantic segmentation on point clouds. Besides, Kundu~\etal~\cite{kundu2020virtual} proposed to fuse features from multiple 2D views for 3D semantic segmentation.
To aggregate together the geometrically-homogeneous points, Landrieu\etal~\cite{landrieu2018large} modeled a point cloud as a super point graph. In addition, a number of recent research works~\cite{zhao2021point,dong2022learning,wu2022point,lai2022stratified} are further proposed to improve the performance of point cloud semantic segmentation. Inspired by~\cite{landrieu2018large}, we expand the sparse labels to geometrically homogeneous super-voxels to generate initial pseudo labels for the first-iteration network training.

\textit{Voxel-based networks} take the regular voxel-grids as input instead of the raw data ~\cite{tchapmi2017segcloud,riegler2017octnet,graham2015sparse,su2018splatnet,dai20183dmv}. 
The recently-proposed methods SparseConv~\cite{graham2017submanifold}, MinkowskiNet~\etal~\cite{choy20194d}, and OccuSeg~\etal~\cite{han2020occuseg} are among the representative works in this branch. 
In this paper, we adopt the 3D-UNet architecture described in~\cite{graham2017submanifold} as the backbone architecture due to its high performance and applicability.

\paragraph{Weakly Supervised 3D Semantic Segmentation} 
Although there have been significant advancements in 3D semantic segmentation, the arduous task of point-level annotation greatly limits its practicality. To overcome this challenge, several approaches have been proposed~\cite{guinard2017weakly,mei2019semantic,wang2020weakly}. 
A recent work~\cite{wei2020multi} utilizes the Class Activation Map to generate pseudo point-wise labels from sub-cloud-level annotations. The performance is, however, limited by the lack of localization information.
In addition, Xu~\etal~\cite{xu2020weakly} achieves the performance close to fully supervised with less than 10\% labels.
However, they require the annotations to be uniformly-distributed in the point cloud, which is practically very hard for the annotators to follow. 
Very recently, several of approaches~\cite{hou2021exploring,tian2022vibus,liu2022active,tang2022learning,dong2022rwseg,yu2022data,zhang2021perturbed,lee2023gaia,ren20213d,kweon2022joint,li2022hybridcr,zhang2021weakly,yang2022mil,deng2022superpoint,wu2022dual,wu2022pointmatch,hu2022sqn} are proposed to further enhance the annotation efficiency and performance of weakly supervised 3D semantic segmentation. 

In this works, we propose a new self-training approach with a label propagation module, in which the network training and label propagation are conducted iteratively.
Our approach largely reduces the reliance on the quality of the initial annotation and achieves top performances, compared with existing weakly supervised methods, while using only extremely-sparse annotations.



\paragraph{3D Instance Segmentation}
3D instance segmentation aims to derive instance-level understanding of a 3D scene, going beyond semantic segmentation. While existing works~\cite{hou20193d, yang2019learning, dong2022learning, schult2022mask3d} use point-level annotations for 3D instance segmentation, the laborious and tedious annotation process limits their practical applicability. Recent works~\cite{hou2021exploring, tao2022seggroup, chu2022twist, tang2022learning} focus on weakly-supervised 3D instance segmentation to overcome this challenge. For example, Seg-Group~\cite{tao2022seggroup} proposes a segment grouping network to hierarchically group unlabeled segments into nearby labeled ones for 3D instance segmentation. Tang et al.~\cite{tang2022learning} use semantic and spatial relations to adaptively learn inter-superpoint affinity. TWIST~\cite{chu2022twist} proposes a semi-supervised 3D instance segmentation approach that leverages object-level information to denoise pseudo labels.
In this work, we extend our self-training approach to 3D instance segmentation using the One-Thing-One-Click annotation.

\paragraph{Self-Training for Semantic Segmentation on 2D Images}
Self-training for weakly supervised 2D image understanding has been intensively explored.
To reduce the annotation burden for 2D images, researchers proposed a variety of annotation approaches,~\eg, image-level categories~\cite{qi2016augmented,oh2017exploiting,zhou2018weakly,ahn2018learning}, points~\cite{bearman2016s,laradji2018blobs}, extreme points~\cite{maninis2018deep,papadopoulos2017extreme},  scribbles~\cite{lin2016scribblesup,wang2019boundary,zhang2020weakly}, bounding boxes~\cite{dai2015boxsup}, etc. With the weak supervision, a self-training approach can learn to expand the limited annotations to unknown regions in the domain. Inspired by the previous works in 2D image understanding, we propose a novel self-training framework for weakly supervised 3D scene understanding.

\begin{figure*}
\centering
\includegraphics[width=0.99\textwidth]{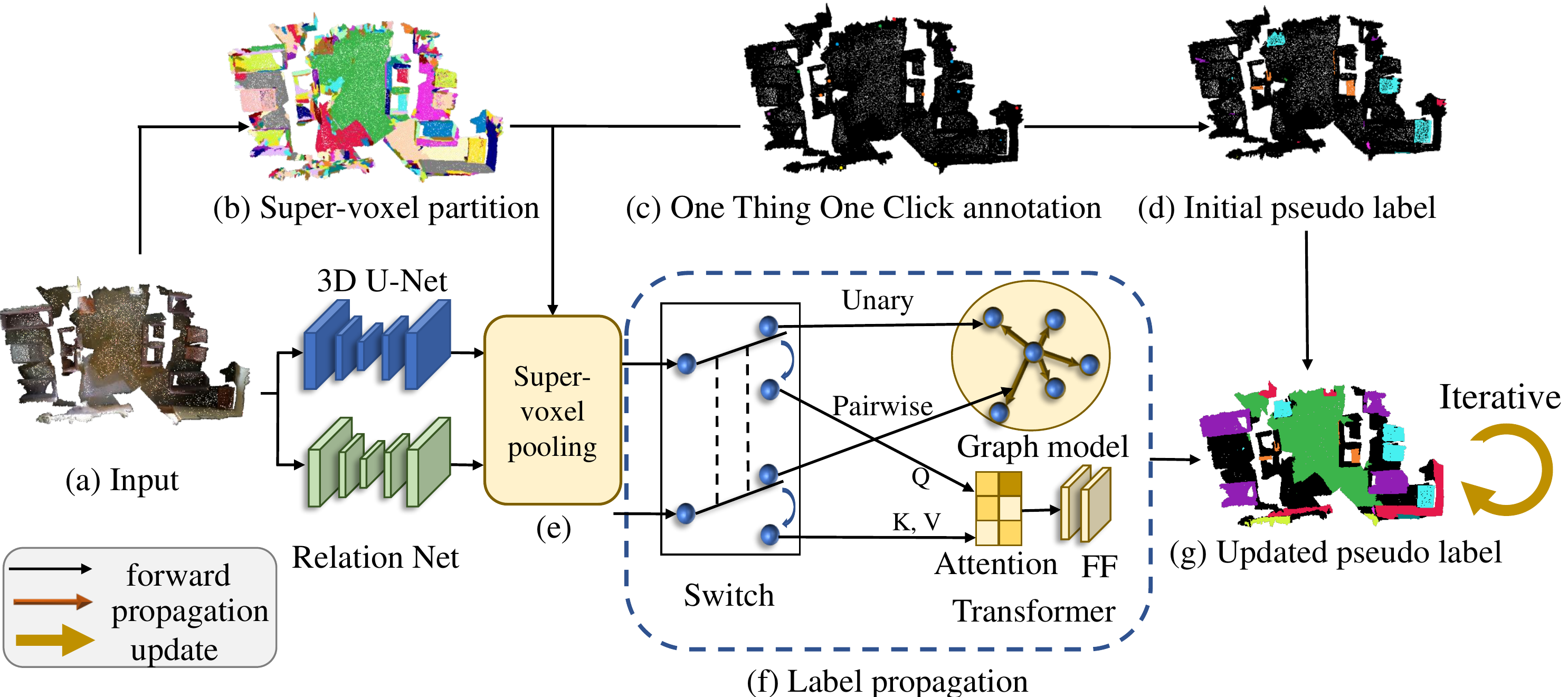}
\caption{Overview of our framework.
Through a super-voxel partition (b), we expand our ``One Thing One Click'' annotations (c) to generate the initial pseudo labels (d) for guiding the update of the pseudo labels (g).
On the other hand, we adopt the ``3D U-Net'' for semantic label prediction (blue region) and design the ``Relation Net'' for super-voxel-based similarity learning (green region).
Then, we incorporate a super-voxel pooling (e) to aggregate features from the two networks. Afterwards, we adopt either a graph model or a transformer (f) to propagate labels over the point cloud.
Further, we iteratively update the predicted labels (g) and train the network.}
\label{fig:overview}

\end{figure*}

\section{Methodology}

\subsection{Overview}

With ``One Thing One Click,'' we only need to annotate a point cloud with one point per object, as Figure~\ref{fig:overview} (c) shows, and these points can be chosen at random to alleviate the annotation burden.  
Procedure-wise, given such sparse annotations, we first over-segment the point cloud $X=\{p_i\}$ into geometrically homogeneous super-voxels $V=\{v_j\}$, where $\cup_j v_j=X$ and $v_j \cap v_{j'}=\emptyset$ for $v_j \neq v_{j'}$.
Note that throughout the paper, we use $i$ and $j$ as the indices for points and super-voxels, respectively.
Based on the super-voxel partition, we can produce initial pseudo labels of the point cloud by spreading each label to all the points locally in the super-voxel that contains the annotated point.
However, as Figure~\ref{fig:overview} (d) shows, the labels are still very sparse. More importantly, the propagated labels distribute mainly around the initially-annotated points, which are far from the ideal uniform distribution for weakly semantic segmentation, as employed in ~\cite{xu2020weakly}.


An important insight in our approach is to iteratively propagate the sparse annotations to unknown regions in the point cloud, while training the network model to guide the propagation process.  To achieve this, we adopt the 3D semantic segmentation network $\Theta$ (the blue regions in Figure~\ref{fig:overview}) and to learn the label propagation via a feature propagation module (Figure~\ref{fig:overview} (f)). Further, we design the relation network $\mathcal{R}$ (the green regions in Figure~\ref{fig:overview}) to explicitly model the feature similarity. Afterward, predictions with high confidence are further employed as the updated pseudo labels for training the network in the next iteration (Figure~\ref{fig:overview} (g)).
This iterative self-training approach couples the label propagation and network training, enabling us to significantly enhance the segmentation quality, as revealed earlier in Figure~\ref{fig:illustration}.



\subsection{3D Semantic Segmentation Network}\label{sec:unary}
We adopt the 3D U-Net architecture~\cite{graham2017submanifold} as the backbone, denoted as $\Theta$. 
Its input is point cloud $X$ of $N$ points (Figure~\ref{fig:overview} (a)).
Each point has 3D coordinates $p_i$ and color $c_i$, where $i\in\{1, ..., N\}$. The network predicts the probability of each semantic category $P(y_{i,\bar{c}}|p_i,c_i,\Theta)$ of each point $p_i$, where $\bar{c}$ is the ground truth category of point $p_i$. The network is trained with the softmax cross-entropy loss below:
%

\begin{equation}
L_{s} = −\frac{1}{N}\sum^N_{i=1}{\log P({y_{i,\bar{c}}|p_i,c_i,\Theta})}
\label{equ:seg_loss}
\end{equation}

\noindent
In the first iteration, the network is trained with the initial pseudo labels, as shown in Figure~\ref{fig:overview} (d).
In subsequent iterations, the network is trained with the updated pseudo labels, as shown in Figure~\ref{fig:overview} (g), which will be detailed below.

\subsection{Pseudo Label Propagation}\label{sec:pairwise}

\begin{figure}
\centering
\includegraphics[width=0.99\columnwidth]{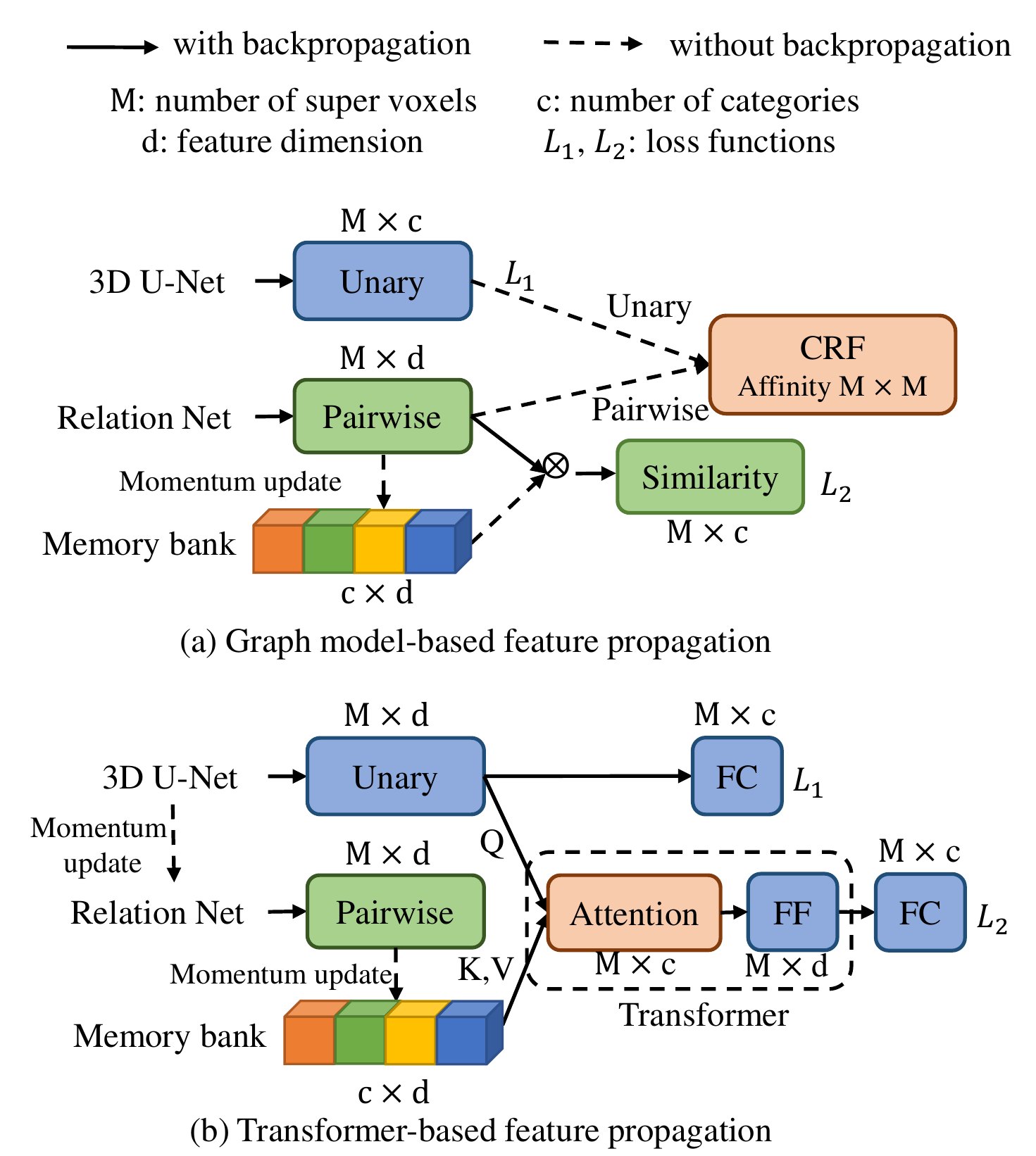}
\caption{The alternative approaches for label propagation. }
\label{fig:relation}
\end{figure}

To facilitate the network training, we propose a label propagation mechanism to effectively propagate labels to unknown regions. Specifically, we provide two options to propagate the feature, \ie, graph-model-based and transformer based, as shown in Figure~\ref{fig:relation}. We also propose the relation network to explicitly learn the similarity among the super-voxels to facilitate the label propagation process and complement the 3D U-Net. 

\paragraph{Graph Model-Based Feature Propagation}

First, we introduce our graph model-based feature propagation. 
To start, we leverage the 3D geometrically homogeneous super-voxels to build a graph.
Compared with building on points, our graph has significant fewer nodes to facilitate efficient label propagation.

To derive the prediction $P(y_{j,c}|v_j,\Theta)$ of the $j$-th super-voxel,
we apply a super-voxel pooling to aggregate the semantic prediction of the $n_j$ points in $v_j$. 

\paragraph{Graph-based Label Propagation}

The architecture of the graph-based label propagation is illustrated in Figure~\ref{fig:relation} (a).
To build the graph, we treat each super-voxel as a graph node and compute the similarity between each pair of super-voxels $v_j,v_{j'}$, which is represented as an edge.

To propagate labels to unknown regions through the graph, we formulate it as an optimization problem that considers both the network prediction and similarities among the super-voxels to achieve the global optimum with the energy function below similar to Conditional Random Field (CRF).
\begin{equation}
\label{equ:energy}
E(Y|V) = \sum_{j} \psi_u (y_{j}|V,\Theta) + \sum_{j<{j'}} \psi_p (y_{j},y_{j'} | V,\mathcal{R},\Theta )
\end{equation}
where $\mathcal{R}$ is the relation network to be detailed later.
The unary term $\psi_u (y_{j}|V,\Theta)$ represents the super-voxel pooled prediction of the 3D U-Net $P(y_j)$ on super-voxel $v_j$. Specifically, it denotes the minus $\log$ probability of predicting super-voxel $v_j$ to have label $y_{j}$. 
We define it as below. 
\begin{equation}
\label{equ:unary_loss}
 \psi_u (y_{j}|V,\Theta) = -\log P({y_{j}|V,\Theta})
\end{equation}

The pairwise term $\psi_p (j_k)$ in Equation~\ref{equ:energy} represents the similarity between super-voxels $v_j$ and $v_{j'}$. We employ both the low-level features and learned features for measuring the similarity, as shown in Equation~\ref{equ:pairwise} below:

\begin{equation}
\begin{aligned}
\label{equ:pairwise}
\psi_p (y_j,y_{j'} | V)= & \mathds{1} (y_j , y_{j'}) \exp\{- \lambda_c \frac{ \left\Vert c_j-c_{j'} \right\Vert^2}{2\sigma^2_c} \\
- \lambda_p \frac{  \left\Vert p_j-p_{j'} \right\Vert^2}{2\sigma^2_p} 
& - \lambda_u \frac{ \left\Vert u_j-u_{j'} \right\Vert^2}{2\sigma^2_u} 
- \lambda_f \frac{ \left\Vert f_j-f_{j'} \right\Vert^2}{2\sigma^2_f} 
\}
\end{aligned}
\end{equation}
where $\mathds{1} (y_j, y_{j'})$ is 1, if $v_j$ and $v_{j'}$ have different predicted labels, and 0 otherwise. The pairwise term means that the cost will be higher if super-voxels with similar features are predicted to be different classes.
Here, $c_j,c_{j'}$, $p_j,p_{j'}$ and $u_j,u_{j'}$ are the normalized mean color, mean coordinates and mean 3D U-Net feature, respectively, of super-voxels $v_j$ and $v_{j'}$.

\paragraph{Relation Network}

Unlike existing works~\cite{zheng2015conditional,chen2017deeplab,yuan2019structpool}, which build the graph on 2D image pixels, we build our graph on 3D super-voxels, which have irregular and complex geometrical structures. Therefore, hand-crafted features $p_j,p_{j'}$ and $c_j,c_{j'}$ have inferior capability to measure the similarity between super-voxels. To address this issue, we propose the \textit{Relation Network} to better leverage the 3D geometrical information and explicitly learn the similarity among super-voxels.



The relation network $\mathcal{R}$ shares the same backbone architecture as the 3D U-Net $\Theta$ except for removing the last category-wise prediction layer. It aims to predict a category-related embedding $f_j$ for each super-voxel $v_j$ as the similarity measurement. $f_j$ is the per super-voxel pooled feature in $\mathcal{R}$. In other words, the relation network groups the embeddings of same category together, while pushing those of different categories apart. To this end, we propose to learn a prototypical embedding for each category. 


To assist the training of the relation network with sparse and unbalanced training data, we present a memory bank $K=\{k\}$ to generate one categorical prototype for each category, instead of simply regarding the average embedding as the prototype as in~\cite{snell2017prototypical}. 

The embedding $f_j$ generated by $\mathcal{R}$ serves as a ``query,'' and we compare it with the corresponding ``key'' $k_c$ in the memory bank with a dot product.
The two modules are optimized simultaneously with contrastive learning~\cite{oord2018representation} as below.

\begin{equation}
\label{equ:contrastive}
L_{c} =\frac{1}{M}\sum^{M}_{j} {(-\log{   \frac{f_j \cdot k_{\bar{c}}/\tau}{\sum_c f_j \cdot k_c/\tau}         })},
\end{equation}

\noindent
where $\tau$ is a temperature hyper parameter~\cite{wu2018unsupervised} and $\bar{c}$ is the ground truth category of $v_j$. The contrastive learning is equivalent to a c-way softmax classification task. 

Following~\cite{he2020momentum}, we update the key representations via a moving average with momentum as shown below
\begin{equation}
\label{equ:momentum}
k_{\bar{c}} \xleftarrow{} m k_{\bar{c}} +(1-m) f_j,
\end{equation}
where $m$ is a momentum coefficient to control the evolving speed of the memory bank. 

Our relation net complements with 3D U-Net. It measures the relations between super-voxels using different training strategies and losses, while 3D U-Net aims to project the inputs into the latent feature space for category assignment. The prediction of relation network is further combined with the prediction of 3D U-Net by multiplying the predicted possibilities of each category to boost the performance. In addition, the relation net offers a stronger measurement of the pairwise term in CRF vs. handcrafted features like colors and also complements with the 3D U-Net features.

\paragraph{Transformer-Based Label Propagation}
In the following, we introduce a transformer-based alternative to the graph model-based approach for label propagation, as illustrated in Figure~\ref{fig:relation} (b). Unlike the graph model-based approach that learns the affinity among super voxels, where the size of the affinity matrix $M\times M$ grows quadratically relative to the number of super voxels $M$, the transformer-based label propagation aims to learn the correlation between a super voxel $v_j$ and a category prototype $k_c$. Therefore, the size of the attention map $M\times c$ grows proportionally to $M$, significantly improving efficiency in terms of memory and inference time. Additionally, transformer-based label propagation can be optimized end-to-end, further improving the performance of 3D semantic segmentation.

Specifically, the transformer-based label propagation can be formulated as follows. 
\begin{equation} 
\hat{f}_{j}=\Sigma_c softmax(\frac{Q(F_j) K(k_c)}{\sqrt{d_l}})V(k_c),
\label{equ:transformer}
\end{equation}
where $Q$, $K$, and $V$ represent MLP layers, while $F_j$ represents the feature vector of the 3D U-Net. The transformer then aggregates the category prototype $k_c$ based on the similarity between $F_j$ and $k_c$. The resulting output feature $\hat{f}_{j}$ is then concatenated with $F_j$ to make the final prediction for the semantic category.

Inspired by Mean Teacher~\cite{tarvainen2017mean}, we update the weights of the Relation Network in our transformer-based label propagation using the moving average of weights in the 3D U-Net with momentum, instead of using stochastic gradient descent (SGD). 
The weight update is formulated as below.  
\begin{equation}
\label{equ:momentum2}
\mathcal{R}_{w} \xleftarrow{} m \mathcal{R}_{w} +(1-m) \Theta_w,
\end{equation}
where $\mathcal{R}{w}$ and $\Theta{w}$ represent the $w$-th weight of the Relation Network and the 3D U-Net, respectively. By using the moving-average strategy, we accumulate the features $F_j$ in the 3D U-Net over time, which improves the quality and stability of the category prototypes $K$. Moreover, this approach helps reduce the computational complexity during training. 

\subsection{Self-Training}\label{sec:self-training}

With the label propagation, we then propose a self-training approach to update networks $\Theta$ and $\mathcal{R}$, and also the pseudo labels $Y$ iteratively. 
The self-training is started by the ``One Thing One Click'' annotations and the pre-constructed super-voxel graph. In each iteration, we fix network parameters $\Theta,\mathcal{R}$ and update label $Y$, and vice versa. There are two steps in each iteration.
\begin{itemize}

\item
With $\Theta$ and $\mathcal{R}$ fixed,
the label propagation is conducted to minimize the energy function in Equation~\ref{equ:energy}.  Then, the predictions with high confidence are taken as the updated pseudo labels for training the two networks in the next iteration.
The confidence of super-voxel $v_j$, denoted as $C_j$, is the average of the minus $\log$ probability of all $n_j$ points in $v_j$ after the label propagation:  

\begin{equation}
\label{equ:momentum3}
C_j = \frac{1}{n_j} \sum_i^{n_j}{ \log P(y_i| p_i, V, \Theta, \mathcal{R}, G)}, \ \text{where} \ p_i \in v_j,
\end{equation}

\noindent
where $G$ denotes the graph propagation. 

\item
With pseudo labels $Y$, $\Theta$ and $\mathcal{R}$ are optimized respectively. 
\end{itemize}

\subsection{3D Instance Segmentation}\label{sec:instance}


To extend our self-training framework for 3D instance segmentation, we propose a point-clustering strategy to iteratively generate instance-level pseudo label and train the instance segmentation network.

\begin{figure*}
\centering
\includegraphics[width=0.99\textwidth]{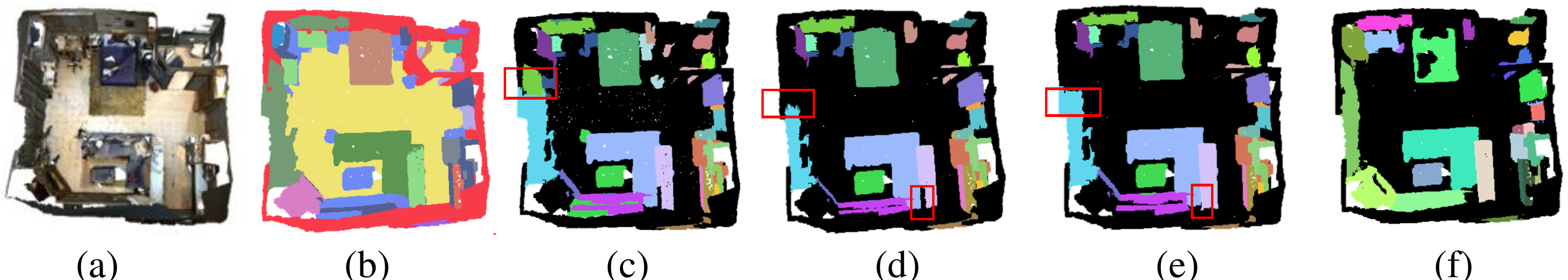}
\caption{Pseudo label generation for 3D instance segmentation. (a) Input point cloud. (b) Our semantic segmentation result. (c) Initial pseudo label by K-Means clustering. (d) Unconnected-segment removal. (e) Results after self-training. (f) Ground truth. }
\label{fig:instance_process}
\end{figure*}

In the first training iteration, we utilize the semantic segmentation network trained with the One Thing One Click annotation approach as described earlier, as shown in Figure~\ref{fig:instance_process} (b). Next, we conduct K-Means Clustering using the annotated super voxels $V_{anno}=\{v_k\}$ as initial centroids. Since our One-Thing-One-Click annotation approach enables each annotated super-voxel to represent an instance, we can predict which instance $k$ the super voxel $v_j$ belongs to based on the Euclidean distance $||p_k-p_j||_2^2$; see Figure~\ref{fig:instance_process} (c). To enhance the robustness of the generated pseudo label, we then filter out small and unconnected semantic segments to obtain the initial instance-level pseudo label (Figure~\ref{fig:instance_process} (d)).

Then we train the instance segmentation network leveraging the above pseudo label. To extend the 3D U-Net architecture for instance-level segmentation, we incorporate a multi-layer perceptron (MLP) prediction head on the top of 3D U-Net following~\cite{jiang2020pointgroup} to predict the 3D offset $o_j$. With the predicted offset, we can move a super voxel $v_j$ towards the centroid coordinate $C$ of the instance that $v_j$ belongs to, such that the super voxel $v_j$ belonging to the same instance can be grouped together and the point-level clustering in the following iterations can be conducted based on the Euclidean distance $||p_k-(p_j+o_j)||_2^2$. 
In the first training iteration, we freeze the backbone network and only update the offset head. This helps the network maintain its ability to perform semantic segmentation.

In the subsequent iterations, we further update the pseudo instance-level label using K-Means Clustering based on $v_j$'s updated coordinate $p_j+o_j$. Then we fine-tune the entire network end-to-end instead of only updating the offset head like the first iteration, so we can further improve the instance segmentation performance.

We repeat the above process and network training iteratively to progressively improve the quality of the pseudo labels and enhance the performance of the model, as illustrated in Figure~\ref{fig:instance_process} (e).



During inference, we use the clustering method from~\cite{jiang2020pointgroup} to group super voxels $v_j$ into candidate clusters based on their predicted shifted coordinates $p_j+o_j$ (P branch in~\cite{jiang2020pointgroup}) instead of their original coordinates $p_j$ (Q branch in~\cite{jiang2020pointgroup}). Additionally, for semantic prediction, we simplify the approach by averaging the predicted semantic scores of all points belonging to each instance, as proposed in~\cite{hou2021exploring}, instead of using the ScoreNet introduced in~\cite{jiang2020pointgroup}.

\section{Experiments}

\paragraph{Datasets}

Our experiments are conducted on two large 3D semantic segmentation datasets -- ScanNet-v2~\cite{dai2017scannet} and S3DIS~\cite{armeni2017joint}.  \textbf{ScanNet-v2}~\cite{dai2017scannet} contains $1513$ 3D scans of 20 semantic categories. We annotate the official training set with our ``One Thing One Click'' scheme, and evaluate on the validation and test set. 
\textbf{S3DIS}~\cite{armeni2017joint} contains 3D scans of $271$ rooms containing $13$ categories. We follow the official train/validation split to annotate on Area 1,2,3,4,6 and report the performance on Area 5. 

\paragraph{``One Thing One Click'' Annotation Details}
In order to ensure the randomness of point selection in annotation, we simulate the annotation procedure by selecting a single point inside an object with the same probability for the following experiments.
In ScanNet-v2, only 19.74 points per scene are annotated on average with ``One Thing One Click'' scheme, 
while this number in the original ScanNet-v2 is 108875.9. In S3DIS, only 36.15 points in each room are annotated on average using ``One Thing One Click'', while the original S3DIS has 193797.1 points annotated in each room. 

\paragraph{Implementation Details}
We implement all the modules of our self-training framework including the mean-field solver~\cite{koller2009probabilistic} for label propagation with the PyTorch~\cite{NEURIPS2019_9015} framework based on the implementation of~\cite{jiang2020pointgroup}. 
Following~\cite{jiang2020pointgroup}, due to the GPU capacity, we randomly choose 250k points if the scene contains more points in training. In inference, the network takes the whole scene as input. We set the hyper-parameters $D=32$, $T=0.9$, $s=20$, $\tau=0.07$, $m=0.9$,
$\sigma_c=\sigma_p=\sigma_u=\sigma_f=1$,  $\lambda_c=\lambda_p=\lambda_u=\lambda_f=1$ with a small validation set.  
We found that the self-training converges after five iterations. After that, more iterations training only brings very minor improvements.

\paragraph{Super-voxel partition}

\begin{figure}
\centering
\includegraphics[width=0.99\columnwidth]{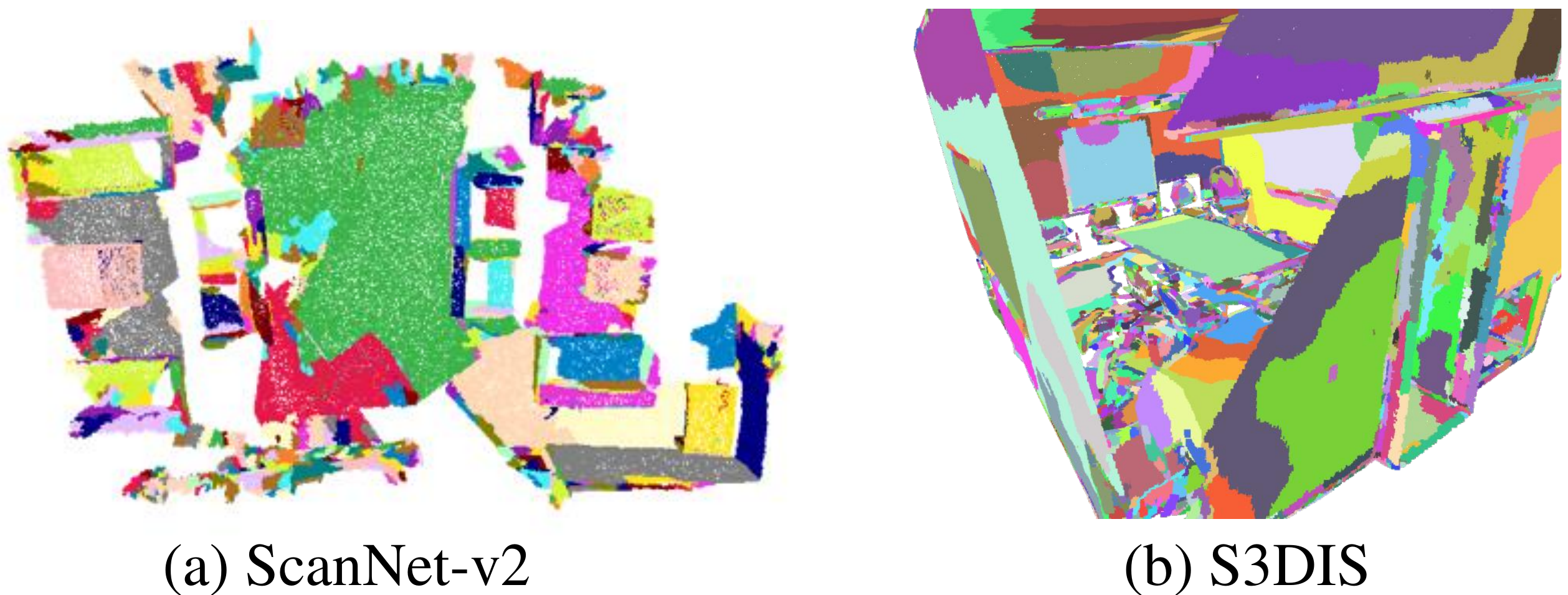}
\caption{Visualization of super-voxel partition on (a) ScanNet-v2 and (b) S3DIS. } \label{fig:supervoxels}
\end{figure}

We use the mesh segment results~\cite{dai2017scannet} as super-voxels for ScanNet-v2, and the geometrical partition results described in~\cite{landrieu2018large} for S3DIS super-voxel partition. Our super-voxel partition effectively groups points based on their geometric attributes, as shown in Figure~\ref{fig:supervoxels}.


\subsection{Semantic Segmentation on ScanNet-v2}\label{sec:scannet}

\paragraph{Comparing with Existing Methods}

Table~\ref{tab:existing} reports the benchmark result on ScanNet-v2 test set. 
The baselines can be roughly divided into two branches.
(i) Fully supervised approaches with 100\% supervision, including several representative works in 3D semantic segmentation.
These methods are the upper bounds of weakly supervised ones.
(ii) Weakly- and semi-supervised approaches. 

With less than 0.02\% annotated points, our result (69.3\% mIoU) outperforms many existing works with full supervision. 
As for weakly- and semi-supervised approaches, our approach outperforms all those very recent ones proposed from 2021 to 2023 with fewer annotations, demonstrating the superiority of our approach over the existing ones. 


\begin{table}
\centering
\scalebox{0.85}{
  \begin{tabular}{c|cc}
    \toprule
    Method & Supervision & mIoU (\%)  \\
    \midrule
    Pointnet++~\cite{qi2017pointnet++} & 100\% &33.9 \\
    SPLATNet~\cite{su2018splatnet}& 100\% & 39.3\\
    TangentConv~\cite{tatarchenko2018tangent} & 100\% &43.8\\
    PointCNN~\cite{li2018pointcnn} & 100\% & 45.8\\
    FPConv~\cite{lin2020fpconv} & 100\% & 63.9\\
    DCM-Net~\cite{schult2020dualconvmesh}&100\%& 65.8 \\
    PointConv~\cite{wu2019pointconv} &100\%& 66.6 \\
    KPConv~\cite{thomas2019kpconv} & 100\% &68.4\\
    JSENet~\cite{hu2020jsenet}& 100\% &69.9 \\
    SubSparseCNN~\cite{graham2017submanifold} & 100\% & 72.5\\
    MinkowskiNet~\cite{choy20194d} &100\% & 73.6 \\
    Virtual MVFusion~\cite{kundu2020virtual} &100\%+2D & 74.6\\
    PointTransformer-v2~\cite{wu2022point} & 100\% & 75.2\\
    Mix3D~\cite{nekrasov2021mix3d} &100\%+2D & 78.1 \\
    \midrule
    Our fully-sup baseline & 100\% & 72.5 \\
    \midrule
    Superpoint-guided~\cite{deng2022superpoint} & 10\% scenes & 52.4\\
    TWIST~\cite{chu2022twist}  & 10\% scenes& 61.1\\ 
    2D Konwledge Transfer~\cite{yu2022data} & 10\% scenes & 61.2 \\
    \midrule
    MPRM~\cite{wei2020multi} & scene-level & 24.4 \\    
    MPRM~\cite{wei2020multi} & subcloud-level & 41.1 \\  
    MPRM+CRF~\cite{wei2020multi} & subcloud-level & 43.2 \\
    WyPR~\cite{ren20213d} & scene-level & 24.0\\
    Zhang~\etal~\cite{zhang2021weakly} & 10.0\% & 52.0   \\
    PSD~\cite{zhang2021perturbed} & 1\% & 54.7 \\
    HybridCR~\cite{li2022hybridcr} & 1\% & 56.8 \\
    SQN~\cite{hu2022sqn} & 0.1\% & 56.9 \\
    SegGroup~\cite{tao2022seggroup} (MinkowskiNet)& 0.028\% & 62.7\\
    GaIA~\cite{lee2023gaia} & 1\% & 65.2 \\
    PointMatch~\cite{wu2022pointmatch} & 0.1\% & 68.8 \\
    One Thing One Click~\cite{liu2021one} & 0.02\% & \textbf{69.1}\\ 
    Ours & 0.02\% & \textbf{69.3}\\ 
    \bottomrule
  \end{tabular}
}
\caption{Comparing with the existing methods on ScanNet-v2 Test Set for 3D Semantic Segmentation.}
\label{tab:existing}
\end{table}

\begin{figure*}
\centering
\includegraphics[width=0.99\textwidth]{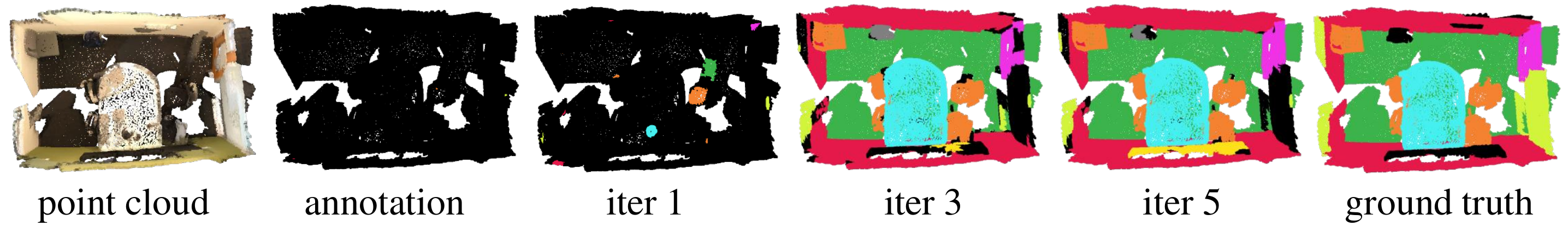}
\caption{Pseudo labels for each iteration on ScanNet-v2 training set. } \label{fig:iter}
\end{figure*}

\paragraph{Results on ScanNet-v2 Data-Efficient Benchmark}
\label{sec:data-efficient}
In Table~\ref{tab:date-efficient}, we show results on ScanNet-v2 ``3D Semantic label with Limited Annotations'' benchmark~\cite{hou2021exploring}. We report the results on the most challenging setting with only 20 points annotated each scene in Table~\ref{tab:baseline} ``Data Efficient''. In this experiment, we use the officially provided 20 points instead of ``One-Thing-One-Click'', and then employ our self-training approach for semantic segmentation. 
The results show that our approach is not limited to ``One Thing One Click'' and is applicable to other annotation schemes. In addition, our approach still outperforms existing works~\cite{hou2021exploring,xie2020pointcontrast,tian2022vibus} under this annotation scheme.

\begin{table}
\centering
\scalebox{0.85}{
  \begin{tabular}{c|cc}
    \toprule
    Method & Supervision & mIoU (\%)  \\
    \midrule
    CSC\_LA\_SEM~\cite{hou2021exploring} & 20 points/scene & 53.1 \\
    PointContrast\_LA\_SEM~\cite{xie2020pointcontrast} & 20 points/scene & 55.0 \\
    VIBUS~\cite{tian2022vibus} & 20 points/scene & 58.6 \\
    \midrule
    One Thing One Click & 20 points/scene & \textbf{59.4}\\ 
    \bottomrule
  \end{tabular}
}
\caption{Comparing with existing methods on ScanNet-v2 Data Efficient Benchmark~\cite{hou2021exploring}.}
\label{tab:date-efficient}
\end{table}

\paragraph{Comparing with Our Baselines}

In this section, we first present three important baselines as shown in Table~\ref{tab:baseline} on ScanNet-v2 validation set. 
\begin{itemize}
\item Table~\ref{tab:baseline} ``Our fully sup baseline'' is trained with the official 100\% annotation provided by ScanNet-v2. It serves as the upper bound of our method. 
\item The model directly trained with the raw annotated points as Figure~\ref{fig:overview} (c) cannot converge well due to the extreme sparsity of the training data.
\item Table~\ref{tab:baseline} ``One Thing One Click$^*$''. The model trained with the initial pseudo labels as Figure~\ref{fig:overview} (d) achieves 62.18\% mIoU. It serves as the starting point of our self-training approach and is denoted as ``our baseline'' in the following. 
\end{itemize}

Table~\ref{tab:baseline} ``One Thing One Click'' manifests that our self-training approach surpasses the baseline by nearly $10\%$ mIoU, attaining a $16\%$ relative improvement. Compared with the fully supervised baseline with the same network architecture, our performance is only 2\% lower. 

Table~\ref{tab:baseline} ``One Thing One Click$^\dagger$'' refers to disabling the graph propagation and relation network in inference. Note that they are still being used in training for generating the pseudo labels. This brings no extra computational burden during the inference, but helps to improve nearly 7\% mIoU, comparing with the baseline (68.96\% vs 62.18\%).

\begin{table}
\centering
\scalebox{0.9}{
  \begin{tabular}{c|ccc}
    \toprule
    Setting & Annotation & mIoU (\%) \\
    \midrule
    Our fully sup baseline & 100\% & 72.18 \\
    \midrule
    One Thing One Click$^*$~\cite{liu2021one} & 0.02\% & 62.18\\
    One Thing One Click$^\dagger$~\cite{liu2021one} & 0.02\% & 68.96\\
    One Thing One Click~\cite{liu2021one} & 0.02\% & \textbf{70.45}\\
     \bottomrule
  \end{tabular}
}
\caption{Our results and baselines on ScanNet-v2 val.~set. $^*$ means the baseline model trained with the initial pseudo labels shown in Figure~\ref{fig:overview} (d). $^\dagger$ means disabling graph propagation and relation network during inference, but note that they are still used in training.  }
\label{tab:baseline}
\end{table}

\paragraph{Results with Fewer Annotations}\label{sec:fewer}

\begin{table}
\centering
\scalebox{0.9}{
  \begin{tabular}{c|cc}
    \toprule
    Method & Annotation (\%) & mIoU (\%) \\
    \midrule
     Two Things One Click$^*$ & 0.01 & 54.71\\
     Two Things One Click$^\dagger$ & 0.01 & 59.56\\ 
     Two Things One Click & 0.01 & \textbf{60.62}\\ 
    \bottomrule
  \end{tabular}
}
\caption{Two Things One Click results and baselines on ScanNet-v2 val.~set. $^*$ means the baseline model trained with the initial pseudo label shown in Figure 3 (d). $^\dagger$ means disabling graph propagation and relation network during inference, but note that they are still used in training. }
\label{tab:2T1C}
\end{table}
To investigate the performance of our approach with even less annotated points, we further annotate ScanNet-v2 with a ``Two Things One Click'' scheme, where we annotate a single random point on half of the objects chosen randomly in the scene. In this way, only less than 0.01\% points are annotated on ScanNet-v2. With the even sparse annotations, we still achieve 60.62\% mIoU as shown in Table~\ref{tab:2T1C}. This experiment also demonstrates that our method can still achieve decent performance even though the annotator ignores several objects by mistake in ``One Thing One Click'' scheme. We further investigate the performance drop with a more challenging ``Four Things One Click'' scheme. However, the model cannot converge well in the very first iteration due to the insufficient label and the self-training fails in this case.

\paragraph{Qualitative Results on ScanNet-v2}
\label{sec:illustration_scannet}

\begin{figure*}
\centering
\includegraphics[width=0.99\textwidth]{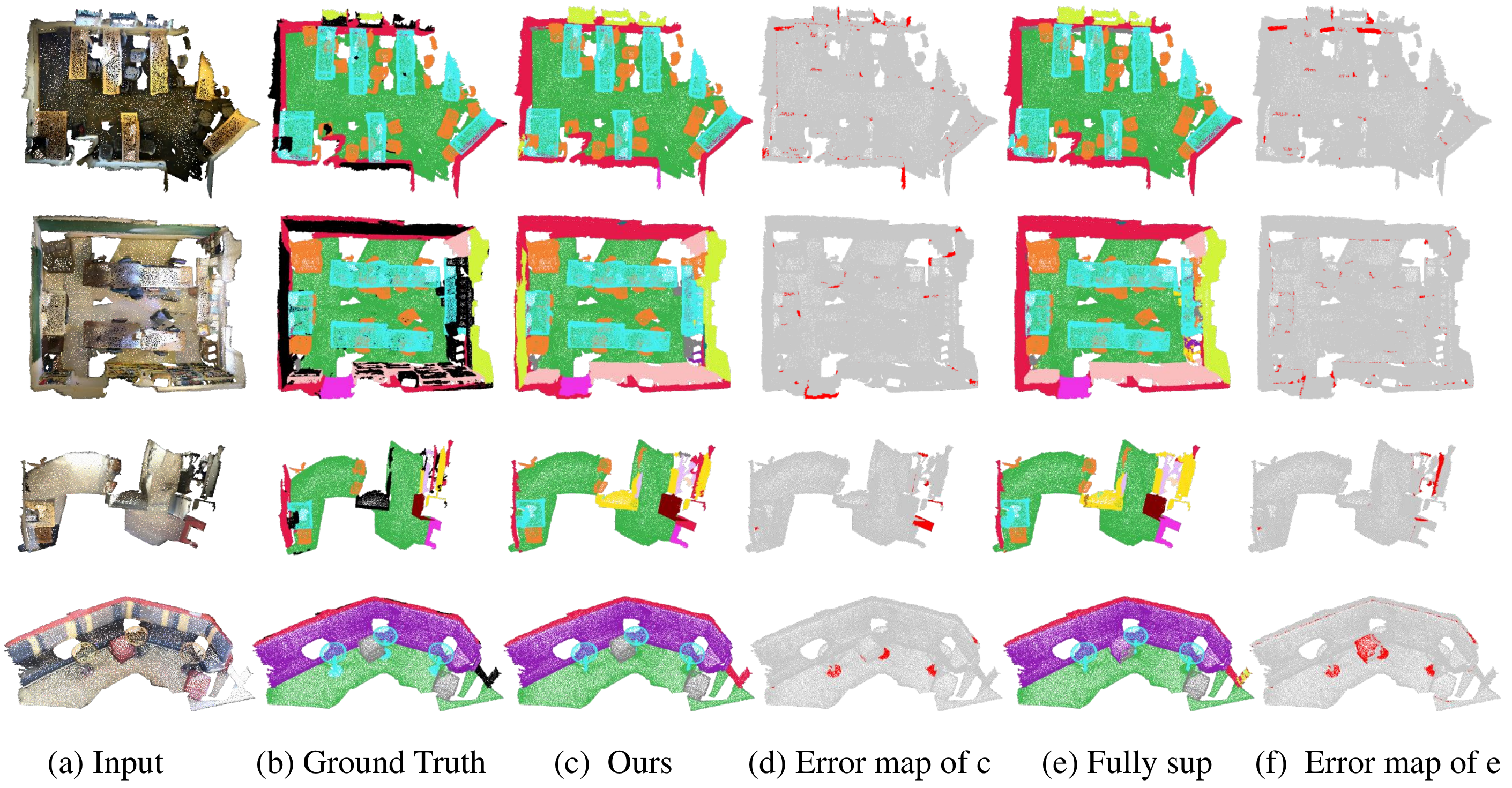}
\caption{Qualitative comparisons on ScanNet-v2. (c) is produced by our model trained only with ``One Thing One Click'' annotations.
(e) is the fully supervised results of~\cite{graham2017submanifold}.
Red regions in (d) and (f) indicate the wrong predictions.
}
\label{fig:scannet_ill}
\end{figure*}

Then, we show prediction results on ScanNet-v2 in Figures~\ref{fig:scannet_ill}.
Through these results, we demonstrate that our approach can produce segmentation results that are comparable to the fully supervised baseline~\cite{graham2017submanifold} with only 0.02\% annotation.
See the error maps shown in (d) and (f) for better visualizations. 

\paragraph{Ablation Studies}

To further study the effectiveness of self-training, graph propagation and relation network, we conduct ablation studies on these three modules on ScanNet-v2 validation set as shown in Table~\ref{tab:ablation} with single view evaluation. 

``3D U-Net'' indicates that the labels are propagated only based on the confidence score of the 3D U-Net itself,~\ie, the unary term in Equation~\ref{equ:energy}. This ablation is designed to manifest the effectiveness of self-training. The ``3D U-Net'' column in Table~\ref{tab:ablation} manifests that the performance is consistently improved with self-training strategy even without pairwise energy term in Equation~\ref{equ:energy} and super-voxel partition. 

``3D U-Net+GP'' refers to the label propagation with graph model, and the similarity among super-voxels are measured by the coordinates $p_i$ and colors $c_i$ without the learned feature $f_i$. This ablation study is to show the effectiveness of the graph model. 
The results in Table~\ref{tab:ablation} indicate that the graph model benefits the label propagation, and finally boosts the overall performance by 2\% over ``3D U-Net'' (67.92\% vs. 65.91\%). 

``3D U-Net+Rel+GP'' utilizes the relation network for similarity measurement based on ``3D U-Net+GP''. In this setting, the similarity among super-voxels is measured with the averaged coordinates $p_i$, the colors $c_i$, the unary features $u_i$, and the relation network generated feature $f_i$, as shown in Equation~\ref{equ:energy}. This experiment is to manifest that the relation network benefits the similarity measurement and pseudo label generation, compared with the hand-crafted feature, i.e., coordinates and color. It outperforms the hand-crafted features especially in the later iterations since the network benefits from the richer pseudo labels. It finally achieves 2.5\% improvement compared with ``3D U-Net+GP'' (70.45\% vs. 67.92\%). 
As shown in Figure~\ref{fig:iter}, the generated pseudo labels for each iteration expands to unknown regions step by step and finally gets close to the ground truth.

\begin{table}
\centering
\scalebox{0.9}{
  \begin{tabular}{c|ccc}
    \toprule
    Method & 3D U-Net & 3D U-Net+GP &3D U-Net+Rel+GP \\
    \midrule
    Iter1 & 60.14  & 63.83 & \textbf{63.92}  \\
    Iter2 & 62.39  & 64.74 & \textbf{66.97} \\
    Iter3 & 64.83  & 66.10 & \textbf{68.40} \\
    Iter4 & 65.81  & 67.78 &  \textbf{70.01} \\
    Iter5 & 65.91  & 67.92 &  \textbf{70.45} \\
    \bottomrule
  \end{tabular}
}
\caption{Ablation studies. ``GP'' indicates the graph propagation, and ``Rel'' means the relation network. ``3D U-Net '' refers to propagating labels only with the network prediction itself.  ``3D U-Net+GP'' indicates label propagation with hand-crafted features. ``3D U-Net+Rel+GP'' indicates label propagation with our relation network. Evaluated on ScanNet-v2 val. set with single view testing. } 
\label{tab:ablation}
\end{table}

\begin{figure}
\centering
\includegraphics[width=0.99\columnwidth]{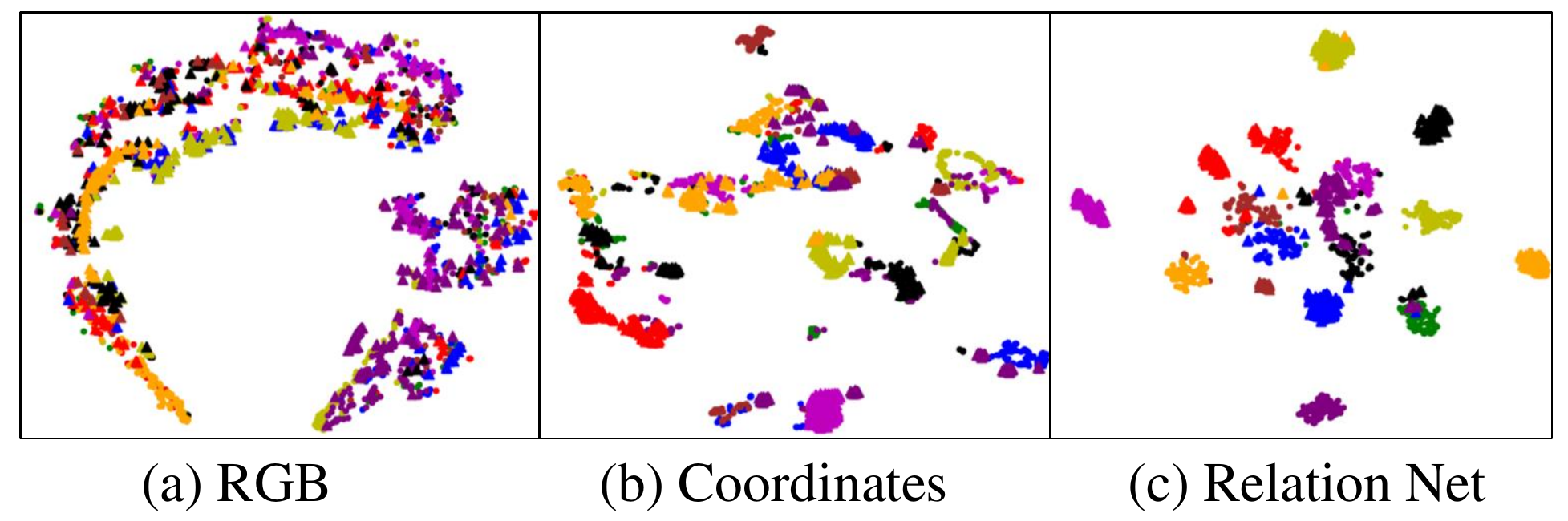}
\caption{The t-SNE visualization of super-voxel features. Different colors and marks (point and plus) indicate different categories. 
The samples of the same category are better grouped together with our relation network (c), compared with hand-crafted features (a \& b). } \label{fig:tsne}
\end{figure}

\begin{table}
\centering
\scalebox{0.9}{
  \begin{tabular}{c|cccc}
    \toprule
    Method & Supervision & AP & AP50 & AP25  \\
    \midrule
    3DSIS~\cite{hou20193d} &100\%  & 16.1 & 38.2 & 55.8\\
    3D-BoNet~\cite{yang2019learning} &100\% & 25.3 & 48.8 & 68.7\\
    RPGN~\cite{dong2022learning} &100\% & 42.8 & 64.3 & 80.6\\
    Mask3D~\cite{schult2022mask3d} &100\% & 56.6 & 78.0 & 87.0\\
    \midrule
    PointGroup~\cite{jiang2020pointgroup} & 100\% & 40.7 & 63.6 & 77.8 \\
    \midrule
    TWIST~\cite{chu2022twist}  & 10\%  & 30.6 & 49.7 &  63.0 \\
    \midrule
    CSC-50~\cite{hou2021exploring} & 0.034\% & 22.9 & 41.4 & 62.0 \\
    SegGroup~\cite{tao2022seggroup} & 0.028\% & 24.6 &44.5 & 63.7\\
    3D-WSIS~\cite{tang2022learning} & 0.02\% & 28.1& 47.2 & \textbf{67.5} \\
    Ours & 0.02\% & \textbf{32.6} &\textbf{52.9}& \textbf{67.5} \\
    \bottomrule
  \end{tabular}
}
\caption{Comparing with existing methods on ScanNet-v2 test set for 3D instance segmentation. PointGroup~\cite{jiang2020pointgroup} means our fully-supervised baseline serving as our upper bound. } 
\label{tab:instance}
\end{table}

\begin{figure*}
\centering
\includegraphics[width=0.89\textwidth]{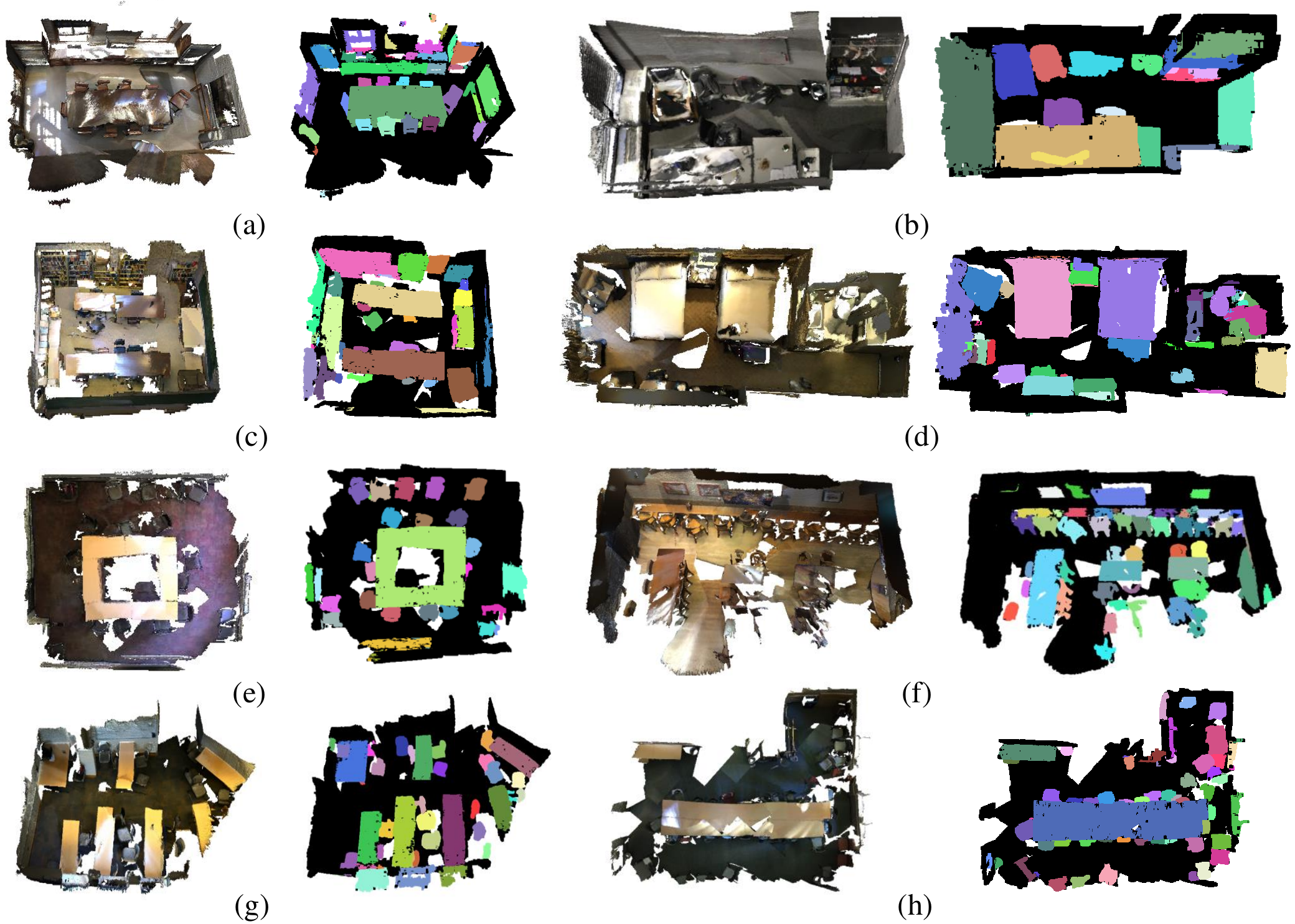}
\caption{Weakly supervised 3D instance segmentation results of our approach. } \label{fig:instance_result}
\end{figure*}

\begin{table}
\centering
\scalebox{0.9}{
  \begin{tabular}{c|ccc}
    \toprule
    Method & Supervision (\%) & mIoU(\%) \\
    \midrule
PointNet~\cite{qi2017pointnet}& 100\% & 41.1 \\
SegCloud~\cite{tchapmi2017segcloud}& 100\% &  48.9 \\
TangentConv~\cite{tatarchenko2018tangent}& 100\% & 52.8\\
3D RNN~\cite{ye20183d}& 100\% & 53.4 \\
PointCNN~\cite{li2018pointcnn}& 100\%& 57.3\\
SuperpointGraph~\cite{landrieu2018large}& 100\%& 58.0\\
MinkowskiNet32~\cite{choy20194d}& 100\% &65.4\\
Virtual MV-Fusion~\cite{kundu2020virtual} & 100\%+2D &65.4\\
    \midrule
    Our fully-sup baseline & 100\%& 63.7 \\
    \midrule
    Superpoint-guided~\cite{deng2022superpoint}& 10\% scenes& 51.1  \\
    \midrule
    $\mathbin{\Pi}$ Model~\cite{laine2016temporal} & 0.2\%&44.3\\
    MT~\cite{tarvainen2017mean}& 0.2\% & 44.4\\
    Xu~\etal
~\cite{xu2020weakly}$^*$ & 0.2\%  &  44.0 \\
    Xu~\etal
~\cite{xu2020weakly} & 0.2\% &44.5 \\  
    $\mathbin{\Pi}$ Model~\cite{laine2016temporal} & 10\%&46.3\\
    MT~\cite{tarvainen2017mean} & 10\% &  47.9\\
   Xu~\etal~\cite{xu2020weakly}$^*$ & 10\% &45.7\\ 
    Xu~\etal~\cite{xu2020weakly} & 10\%  &48.0\\    
    GPFN~\cite{wang2020weakly} & 16.7\% 2D& 50.8 \\  
    GPFN~\cite{wang2020weakly}  & 100\% 2D& 52.5   \\
   Zhang~\etal~\cite{zhang2021weakly} & 0.03\% & 45.8   \\
   Joint 2D-3D~\etal~\cite{kweon2022joint} & Scene+Image & 47.4   \\
   MulPro~\cite{su2022weakly} & 10\% & 49.0 \\
    MIL transformer\cite{yang2022mil}  &  0.02\% & 51.4  \\
    HybridCR~\cite{li2022hybridcr} & 0.03\% & 51.5\\
    GaIA~\cite{lee2023gaia} & 0.02\% & 53.7 \\
    DAT~\cite{wu2022dual} & 0.02\% & 54.6  \\
    \midrule
   One Thing One Click~\cite{liu2021one} & 0.02\% & 50.1  \\ 
    Ours & 0.02\% &\textbf{56.6} \\ 
    \bottomrule
  \end{tabular}
}
\caption{Comparison with existing methods and baselines on 
the S3DIS Area-5. }
\label{tab:s3dis}
\end{table}

\begin{figure*}
\centering
\includegraphics[width=0.99\textwidth]{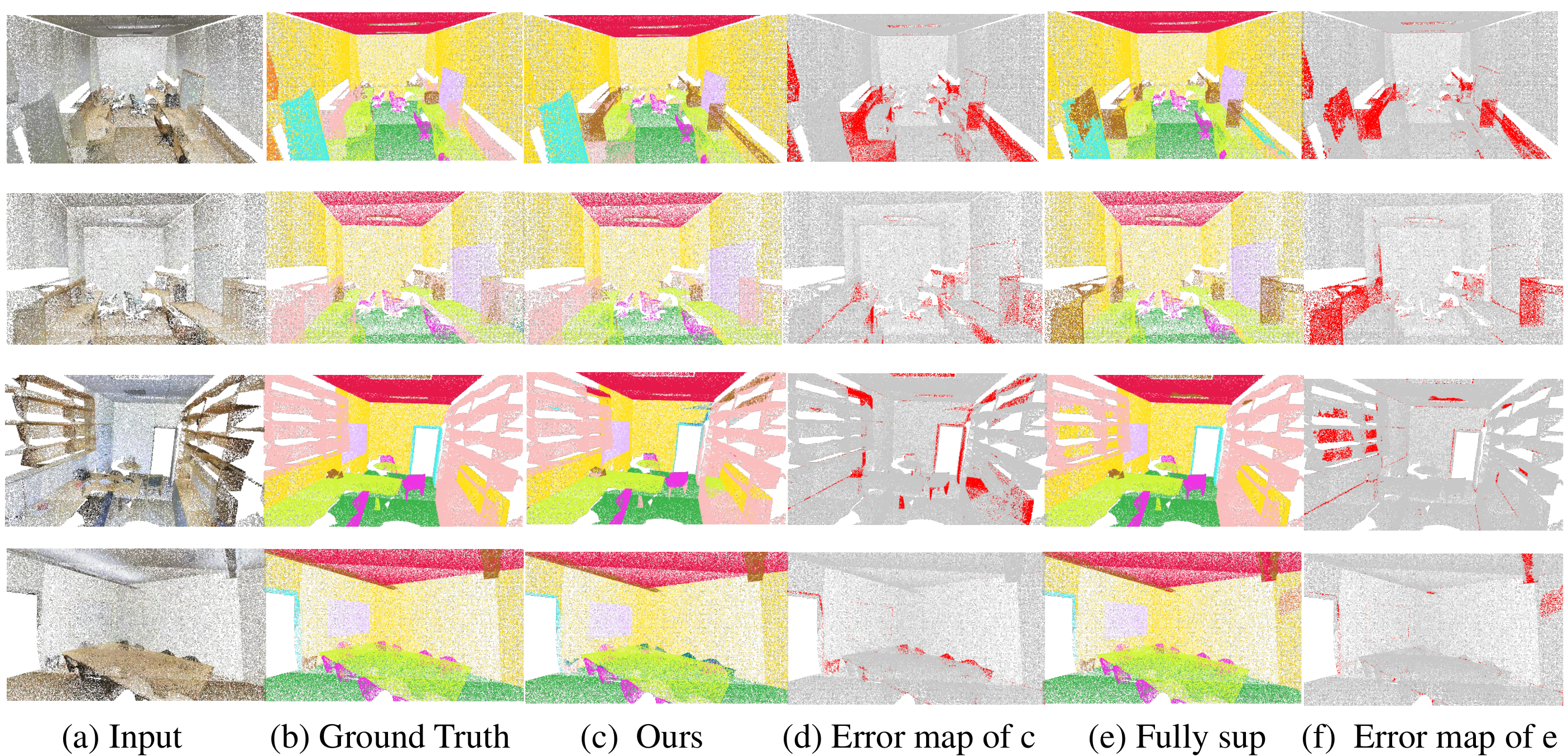}
\caption{Qualitative results on S3DIS. (c) is produced by our model trained only with ``One Thing One Click'' annotations.
(e) is the fully supervised results of~\cite{graham2017submanifold}.
Red regions in (d) and (f) indicate the wrong predictions.
}
\label{fig:s3dis_ill}

\end{figure*}

\paragraph{Analysis of Relation Network}

Further, we study whether the learned embeddings of the relation network outperform the hand-crafted features for similarity measurement. We randomly sample 200 super-voxels for each category in ScanNet-v2, and conduct a t-SNE visualization~\cite{maaten2008visualizing} on them. Figure~\ref{fig:tsne} indicates that the relation network better groups the intra-class embeddings and distinguish the inter-class embeddings compared with hand-crafted features.

\subsection{Instance Segmentation on ScanNet-v2}\label{sec:instance}

Figure~\ref{fig:instance_result} and Table~\ref{tab:instance} show the qualitative and quantitative evaluation results for instance segmentation. 
Figure~\ref{fig:instance_result} demonstrates the effectiveness of our approach in recognizing nearby chairs as individual instances as shown in (a), (e), (f), and (h). Also, our approach can accurately predict larger instances like tables and beds as a single entity; see (d), (e), and (h). Further, the results in Table~\ref{tab:instance} indicate that our approach outperforms all existing methods and even surpasses some fully-supervised approaches such as~\cite{hou20193d,yang2019learning} for 3D instance segmentation.

\subsection{Evaluations on S3DIS}
We also evaluate our approach on the S3DIS dataset. Only less than 0.02\% points in the dataset are annotated with our ``One Thing One Click'' scheme. 

\paragraph{Comparing with Existing Works}

We also compare with fully supervised approaches and weakly supervised approaches on S3DIS.
As shown in Table~\ref{tab:s3dis}, with the ``One Thing One Click'' scheme where less than 0.02\% points are annotated, we achieve 56.6\% mIoU, outperforming existing works by a considerable margin, including our previous version~\cite{liu2021one} (50.1\%).  


In addition, our approach achieves comparable results with several fully supervised methods as shown in Table~\ref{tab:s3dis}.

\vspace{-0.1in}

\paragraph{Qualitative Results on S3DIS}
\label{sec:illustration_s3dis}

Figures~\ref{fig:s3dis_ill} illustrates our results on S3DIS. Again, with only 0.02\% annotation, our approach can produce high quality semantic predictions (c) that are comparable to the fully-supervised approach (e). See the error maps (d, f) for better illustration. 

\section{Limitations}

Despite of the good performance, there are still several limitations of our approach. For instance, our approach brings extra training burden compared with our fully-supervised baseline~\cite{graham2017submanifold,graham20183d} . In the model training, our approach requires additional training epochs and also optimizes a relation network besides the 3D U-Net. Note that we mainly consider the 3D U-Net and relation network since the computational burden of other modules can be neglected.  
Specifically, in the first training iteration, we train the network for 512 epochs following~\cite{graham2017submanifold,graham20183d} , and then fine-tune the network for 256 epochs for the remaining iterations; hence, we need three times of standard training time $T$ of the baseline model, denoted as $T^\prime=3T$. In addition, optimizing the relation network (taking the transformer-based option as an example) requires the same number of forward passes as 3D U-Net without back-propagation, and the training time is roughly $T^\prime/2=3T/2$. Therefore, the overall training time of our approach is roughly $4.5T$. Besides, our approach doubles the number of parameters due to the relation network. Fortunately, in inference, the feed-forward pass of 3D U-Net and relation network can be conducted in parallel to keep the inference time roughly equal to our fully-supervised baseline~\cite{graham2017submanifold,graham20183d} .

\section{Conclusion}

We propose the ``One Thing One Click'' scheme to efficiently annotate point clouds for weakly supervised 3D semantic segmentation, requiring significantly fewer annotations than the previous approaches.
%
To put this scheme into practice, we formulate a self-training approach to make it feasible for the network to learn from such extremely sparse labels.
Specifically, we execute the two key modules in our approach iteratively: expand labels through the label propagation module and train the network using the updated pseudo labels.
Further, we adopt a relation network to explicitly learn the feature similarity. In addition, our approach is compatible to 3D instance segmentation using the One Thing One Click annotations. 
Experiments on two large 3D datasets ScanNet-v2 and S3DIS manifest that our approach, with only extremely-sparse annotations, outperforms existing weakly supervised methods on 3D semantic segmentation and instance segmentation consistently. Moreover, our results are even comparable to those of the fully supervised counterparts.

{\small
\bibliographystyle{ieee_fullname}
\bibliography{cvpr}
}

\end{document}